# Memory Capacity of Neural Networks using a Circulant Weight Matrix

Vamsi Sashank Kotagiri
Oklahoma State University, Stillwater

**Abstract**
This paper presents results on the memory capacity of a generalized feedback neural network using a circulant matrix. Children are capable of learning soon after birth which indicates that the neural networks of the brain have prior learnt capacity that is a consequence of the regular structures in the brain's organization. Motivated by this idea, we consider the capacity of circulant matrices as weight matrices in a feedback network.

**Introduction**
Children are capable of learning soon after birth [1]-[4] which indicates that the neural networks of the brain have prior learnt capacity that is a consequence of the regular structures in the brain's organization. The simplest regularity that can be conceived is that of a circulant structure. Therefore, we consider the memory storage behavior and capacity of feedback networks that have circulant structure in the weight matrix.

The mathematical background to the use of circulant matrices is provided in [5]-[10]. Most recently these ideas have been applied to the development of number theoretic transforms [11]-[13]. The long term objective is to see if such circulant matrix based processing might have applications also to larger questions underlying the storage of memories [14]-[19]. These include non-classical processing [15]-[19] and abstract concept formation [20]-[30]. Our approach is basically Hebbian [31],[32]. We show that these networks can store substantial number of patterns and their shifted versions. The memory capacity of even sized networks is higher than their nearly situated odd sized networks.

**Background**
A Hopfield network is a fully connected feed-back neural network in which the connections between pair of neurons is characterized by weights and an N×N symmetric matrix which is build by using weights. This is the simplest and most widely used network model. A Hopfield network consists of artificial neurons which have N inputs. With each input i there will be a weight $w_i$ and also have an output state. The state of the output can be maintained until the neuron is updated. Previous results on the capacity of Hopfield networks are given in [32]-40].

The following operations help us in updating these neurons:

    The value of each input $x_i$ is determined and the sum of all weighted inputs $\sum w_i x_i$ is calculated.

The output state of the neuron is set to +1 if the weighted input sum is larger or equal to 0. It is set to -1 if the weighted input sum is smaller than 0.

When written as a formula:

$$o = \begin{cases} 1 : \sum w_i x_i \geq 0 \\ -1 : \sum w_i x_i < 0 \end{cases}$$

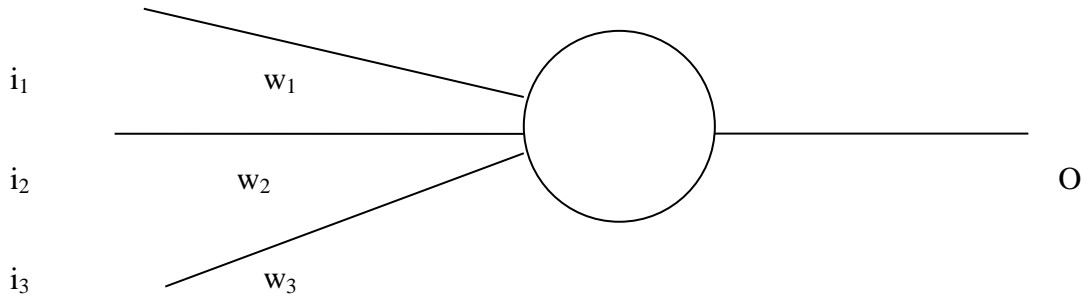

Figure 1: Artificial neuron used in Hopfield network

The weight matrix W is a N×N symmetric matrix whose components $w_{ij}$ and $w_{ji}$ are the same. In other words, the matrix is a symmetric matrix so $w_{ij} = w_{ji}$.

$$\begin{bmatrix} 0 & w_{12} & w_{13} & w_{14} & w_{15} \\ w_{21} & 0 & w_{23} & w_{24} & w_{25} \\ w_{31} & w_{32} & 0 & w_{34} & w_{35} \\ w_{41} & w_{42} & w_{43} & 0 & w_{45} \\ w_{51} & w_{52} & w_{53} & w_{54} & 0 \end{bmatrix}$$

Using the Hopfield network as a reference we are going to use N×N circulant matrix and try to store a set of patterns as neural memories which are relevant to the circulant matrix chosen.

The circulant matrix is as follows

$$C = \begin{bmatrix} 0 & a & 0 & b & 0 & c & 0 & . & . & k \\ k & 0 & a & 0 & b & 0 & c & 0 & . & . \\ . & k & 0 & a & 0 & b & 0 & c & 0 & . \\ . & . & k & 0 & a & 0 & b & 0 & c & 0 \\ 0 & . & . & k & 0 & a & 0 & b & 0 & c \\ c & 0 & . & . & k & 0 & a & 0 & b & 0 \\ 0 & c & 0 & . & . & k & 0 & a & 0 & b \\ b & 0 & c & 0 & . & . & k & 0 & a & 0 \\ 0 & b & 0 & c & 0 & . & . & k & 0 & a \\ a & 0 & b & 0 & c & 0 & . & . & k & 0 \end{bmatrix}$$

**4×4 circulant matrix**:

Consider a 4×4 circulant matrix which consists of 0, a, b, c as elements. Some random value can be assigned to each of the variable and the matrix looks as follows. We can divide these circulant matrices which hold memories namely as two classes (based on the sum of the elements used in the matrix).

Class-1:
    In this class the sum of the elements in the matrix is equal to zero.

$$\begin{bmatrix} 0 & 2 & -5 & 3 \\ 3 & 0 & 2 & -5 \\ -5 & 3 & 0 & 2 \\ 2 & -5 & 3 & 0 \end{bmatrix}$$

$$\begin{bmatrix} 0 & 2 & -7 & 5 \\ 5 & 0 & 2 & -7 \\ -7 & 5 & 0 & 2 \\ 2 & -7 & 5 & 0 \end{bmatrix} \begin{bmatrix} 0 & 5 & -8 & 3 \\ 3 & 0 & 5 & -8 \\ -8 & 3 & 0 & 5 \\ 5 & -8 & 3 & 0 \end{bmatrix} \begin{bmatrix} 0 & 1 & -4 & 3 \\ 3 & 0 & 1 & -4 \\ -4 & 3 & 0 & 1 \\ 1 & -4 & 3 & 0 \end{bmatrix}$$

The above matrices are circulant in nature and were randomly generated. By observation we can make a comment that a 4×4 matrix with +1's and -1's can make out total memory outcomes as $2^4=16$. Out of these total available memories by conducting experiments the circulant matrix of this form (where the sum of elements used in the matrix is zero) holds 5 memories and of those 5 memories held by the matrix 3 of them are unique memories by not counting the simple inversions. Unless or until the pattern mentioned in the above example matrices and sum of elements used in the circulant matrix is equal to zero then that circulant matrix will hold only these horizontal memories.

$$\begin{matrix}[+ & + & - & -]\\ [- & + & + & -]\\ [- & - & + & +]\\ [+ & - & - & +]\\ [+ & + & + & +]\end{matrix}$$

Class-2:
  The sum of the elements selected for the circulant matrix is either 1 or 2.

$$\begin{bmatrix} 0 & 2 & -5 & 4 \\ 4 & 0 & 2 & -5 \\ -5 & 4 & 0 & 2 \\ 2 & -5 & 4 & 0 \end{bmatrix} \begin{bmatrix} 0 & 5 & -7 & 3 \\ 3 & 0 & 5 & -7 \\ -7 & 3 & 0 & 5 \\ 5 & -7 & 3 & 0 \end{bmatrix} \text{ when sum of elements is 1}$$

$$\begin{bmatrix} 0 & 5 & -6 & 3 \\ 3 & 0 & 5 & -6 \\ -6 & 3 & 0 & 5 \\ 5 & -6 & 3 & 0 \end{bmatrix} \text{ when sum of elements is 2.}$$

Out of these total available memories by conducting experiments the circulant matrix of this form (where the sum of elements used in the matrix is either 1 or 2) holds 6 memories and of those 6 memories held by the matrix 3 of them are unique memories by not counting the simple inversions. Unless or until the pattern mentioned in the above example matrices and sum of elements used in the circulant matrix is equal to 1 or 2 then that circulant matrix will hold only these vertical memories

$$\begin{matrix}[+ & + & - & -]\\ [- & + & + & -]\\ [- & - & + & +]\\ [+ & - & - & +]\\ [- & - & - & -]\\ [+ & + & + & +]\end{matrix}$$

**5×5 circulant matrix**:

Let us consider a 5×5 matrix whose elements are 0, a, b, c, d which are represented in a circulant nature.

$$\begin{bmatrix} 0 & -2 & 3 & 3 & -2 \\ -2 & 0 & -2 & 3 & 3 \\ 3 & -2 & 0 & -2 & 3 \\ 3 & 3 & -2 & 0 & -2 \\ -2 & 3 & 3 & -2 & 0 \end{bmatrix}$$

The matrix is 5×5 so the total outcomes of memory capacity are $2^5=32$. The 5×5 matrix can hold 7 memories out of available memories and of those it can hold 6 unique memories, not counting the simple inversions.

$$\begin{bmatrix} 0 & -1 & 2 & 2 & -1 \\ -1 & 0 & -1 & 2 & 2 \\ 2 & -1 & 0 & -1 & 2 \\ 2 & 2 & -1 & 0 & -1 \\ -1 & 2 & 2 & -1 & 0 \end{bmatrix} \begin{bmatrix} 0 & -3 & 4 & 4 & -3 \\ -3 & 0 & -3 & 4 & 4 \\ 4 & -3 & 0 & -3 & 4 \\ 4 & 4 & -3 & 0 & -3 \\ -3 & 4 & 4 & -3 & 0 \end{bmatrix}$$

$$\begin{bmatrix} + & + & + & + & + \\ + & + & - & + & - \\ + & - & + & + & - \\ - & + & - & + & + \\ - & + & + & - & + \\ + & - & + & - & + \\ - & - & - & - & - \end{bmatrix}$$

$$\begin{bmatrix} 0 & 3 & -1 & -1 & 3 \\ 3 & 0 & 3 & -1 & -1 \\ -1 & 3 & 0 & 3 & -1 \\ -1 & -1 & 3 & 0 & 3 \\ 3 & -1 & -1 & 3 & 0 \end{bmatrix}$$

$$\begin{bmatrix} + & + & + & + & + \\ + & - & - & + & + \\ + & + & - & - & + \\ + & + & + & - & - \\ - & + & + & + & - \\ - & - & + & + & + \\ - & - & - & - & - \end{bmatrix}$$

**6×6 circulant matrix**:
Let us consider a 6×6 matrix whose elements are 0, a, b, c, d, e which are represented in a circulant nature.

$$\begin{bmatrix} 0 & -4 & 1 & 2 & 3 & -2 \\ -2 & 0 & -4 & 1 & 2 & 3 \\ 3 & -2 & 0 & -4 & 1 & 2 \\ 2 & 3 & -2 & 0 & -4 & 1 \\ 1 & 2 & 3 & -2 & 0 & -4 \\ -4 & 1 & 2 & 3 & -2 & 0 \end{bmatrix}$$

The matrix is 6×6 so the total outcomes of memory capacity are $2^6$=64. The 6×6 matrix can hold 9 memories out of available memories and of those it can hold 5 unique memories, not counting the simple inversions.

[+  +  +  +  +  +]
[−  +  −  −  +  −]
[+  −  +  +  −  +]
[+  +  −  +  +  −]
[−  −  +  −  −  +]
[−  +  +  −  +  +]
[+  −  −  +  −  −]
[+  −  +  −  +  −]
[−  +  −  +  −  +]

**7×7 circulant matrix**:
Let us consider a 7×7 matrix whose elements are 0, a, b, c, d, e, f which are represented in a circulant nature.

$$\begin{bmatrix} 0 & -2 & -1 & 3 & 3 & 1 & -2 \\ -2 & 0 & -2 & -1 & 3 & 3 & 1 \\ 1 & -2 & 0 & -2 & -1 & 3 & 3 \\ 3 & 1 & -2 & 0 & -2 & -1 & 3 \\ 3 & 3 & 1 & -2 & 0 & -2 & -1 \\ -1 & 3 & 3 & 1 & -2 & 0 & -2 \\ -2 & -1 & 3 & 3 & 1 & -2 & 0 \end{bmatrix}$$

The matrix is 7×7 so the total outcomes of memory capacity are $2^7$=128. The 7×7 matrix can hold 9 memories out of available memories and of those it can hold 8 unique memories, not counting the simple inversions.

[−  −  −  −  −  −  −]
[+  +  +  −  +  +  −]
[+  +  −  +  +  +  −]
[+  −  +  +  +  −  +]
[+  +  −  +  +  −  +]
[+  −  +  +  −  +  +]
[+  +  +  +  +  +  +]

$$\begin{matrix}[- & + & + & - & + & + & +]\\ [- & + & + & + & - & + & +]\end{matrix}$$

**8 ×8 circulant matrix**:

Let us consider an 8×8 matrix whose elements are 0, a, b, c, d, e, f, g which are represented in a circulant nature. We can divide these circulant matrices which hold memories namely as two classes (based on the sum of the elements used in the matrix).

Class -1:

In this class the sum of the elements in the matrix is equal to zero

$$\begin{bmatrix} 0 & -2 & -1 & 4 & 4 & 1 & -2 & -4\\ -4 & 0 & -2 & -1 & 4 & 4 & 1 & -2\\ -2 & -4 & 0 & -2 & -1 & 4 & 4 & 1\\ 1 & -2 & -4 & 0 & -2 & -1 & 4 & 4\\ 4 & 1 & -2 & -4 & 0 & -2 & -1 & 4\\ 4 & 4 & 1 & -2 & -4 & 0 & -2 & -1\\ -1 & 4 & 4 & 1 & -2 & -4 & 0 & -2\\ -2 & -1 & 4 & 4 & 1 & -2 & -4 & 0 \end{bmatrix}$$

By observation we can make a comment that above matrix is circulant in nature and outcome matrix can be generated by using +1's and -1's which sum up value as $2^8$=256. Out of these the 8×8 matrix can hold 11 memories and of those 11 memories held by the matrix 10 of them are unique memories by not counting the simple inversions.

$$\begin{matrix}[+ & + & + & - & + & + & + & -]\\ [- & + & + & + & - & + & + & +]\\ [+ & - & + & + & + & - & + & +]\\ [+ & + & - & + & + & + & - & +]\\ [+ & - & + & - & + & - & + & -]\\ [- & + & - & + & - & + & - & +]\\ [+ & + & - & - & + & + & - & -]\\ [- & + & + & - & - & + & + & -]\\ [- & - & + & + & - & - & + & +]\\ [+ & - & - & + & + & - & - & +]\\ [+ & + & + & + & + & + & + & +]\end{matrix}$$

Class-2:

The sum of the elements selected for the circulant matrix is 1.

$$\begin{bmatrix} 0 & -2 & -1 & 4 & 5 & 1 & -2 & -4 \\ -4 & 0 & -2 & -1 & 4 & 5 & 1 & -2 \\ -2 & -4 & 0 & -2 & -1 & 4 & 5 & 1 \\ 1 & -2 & -4 & 0 & -2 & -1 & 4 & 5 \\ 5 & 1 & -2 & -4 & 0 & -2 & -1 & 4 \\ 4 & 5 & 1 & -2 & -4 & 0 & -2 & -1 \\ -1 & 4 & 5 & 1 & -2 & -4 & 0 & -2 \\ -2 & -1 & 4 & 5 & 1 & -2 & -4 & 0 \end{bmatrix}$$

Here we can observe that in the above circulant matrix the sum of elements is 1 and by conducting random experiments on this circulant matrix, a conclusion can be drawn from those experiments that the above circulant matrix holds 16 memories out of the available memories and of those 16 memories held by the matrix, 10 of them are unique memories by not counting the simple inversions.

$$\begin{matrix}
[+ & + & + & - & + & + & + & -] \\
[- & + & + & + & - & + & + & +] \\
[+ & - & + & + & + & - & + & +] \\
[+ & + & - & + & + & + & - & +] \\
[+ & - & + & - & + & - & + & -] \\
[- & + & - & + & - & + & - & +] \\
[+ & + & - & - & + & + & - & -] \\
[- & + & + & - & - & + & + & -] \\
[- & - & + & + & - & - & + & +] \\
[+ & - & - & + & + & - & - & +] \\
[- & - & - & + & - & - & - & +] \\
[+ & - & - & - & + & - & - & -] \\
[- & + & - & - & - & + & - & -] \\
[- & - & + & - & - & - & + & -] \\
[- & - & - & - & - & - & - & -] \\
[+ & + & + & + & + & + & + & +]
\end{matrix}$$

### 9 ×9 circulant matrix:

Let us consider a 9×9 matrix whose elements are 0, a, b, c, d, e, f, g, h which are represented in a circulant nature.

$$\begin{bmatrix} 0 & -3 & -2 & -1 & 4 & 4 & 1 & 2 & -3 \\ -3 & 0 & -3 & -2 & -1 & 4 & 4 & 1 & 2 \\ 2 & -3 & 0 & -3 & -2 & -1 & 4 & 4 & 1 \\ 1 & 2 & -3 & 0 & -3 & -2 & -1 & 4 & 4 \\ 4 & 1 & 2 & -3 & 0 & -3 & -2 & -1 & 4 \\ 4 & 4 & 1 & 2 & -3 & 0 & -3 & -2 & -1 \\ -1 & 4 & 4 & 1 & 2 & -3 & 0 & -3 & -2 \\ -2 & -1 & 4 & 4 & 1 & 2 & -3 & 0 & -3 \\ -3 & -2 & -1 & 4 & 4 & 1 & 2 & -3 & 0 \end{bmatrix}$$

The above circulant matrix is 9×9 so the total outcomes of memory capacity are $2^9$=512. The 9×9 matrix can hold 11 memories out of available memories and of those it can hold 9 unique memories, not counting the simple inversions.

$$\begin{matrix}
[- & - & - & - & - & - & - & - & -] \\
[+ & + & + & + & - & + & + & + & -] \\
[- & + & + & + & + & - & + & + & +] \\
[+ & - & + & + & + & + & - & + & +] \\
[+ & + & - & + & + & + & + & - & +] \\
[+ & + & + & - & + & + & + & + & -] \\
[- & + & + & + & - & + & + & + & +] \\
[+ & - & + & + & + & - & + & + & +] \\
[+ & + & - & + & + & + & - & + & +] \\
[+ & + & + & - & + & + & + & - & +] \\
[+ & + & + & + & + & + & + & + & +]
\end{matrix}$$

**10 ×10 circulant matrix**:
Let us consider a 10×10 matrix whose elements are 0, a, b, c, d, e, f, g, h, i which are represented in a circulant nature.

$$\begin{bmatrix}
0 & -4 & -1 & -2 & 3 & 11 & 2 & -1 & -3 & -4 \\
-4 & 0 & -4 & -1 & -2 & 3 & 11 & 2 & -1 & -3 \\
-3 & -4 & 0 & -4 & -1 & -2 & 3 & 11 & 2 & -1 \\
-1 & -3 & -4 & 0 & -4 & -1 & -2 & 3 & 11 & 2 \\
2 & -1 & -3 & -4 & 0 & -4 & -1 & -2 & 3 & 11 \\
11 & 2 & -1 & -3 & -4 & 0 & -4 & -1 & -2 & 3 \\
3 & 11 & 2 & -1 & -3 & -4 & 0 & -4 & -1 & -2 \\
-2 & 3 & 11 & 2 & -1 & -3 & -4 & 0 & -4 & -1 \\
-1 & -2 & 3 & 11 & 2 & -1 & -3 & -4 & 0 & -4 \\
-4 & -1 & -2 & 3 & 11 & 2 & -1 & -3 & -4 & 0
\end{bmatrix}$$

The above circulant matrix is 10×10 so the total outcomes of memory capacity are $2^{10}$=1024.The 10×10 matrix can hold 24 memories out of available memories and of those it can hold 21 unique memories, not counting the simple inversions.

$$\begin{matrix}
[+ & + & + & + & - & + & + & + & + & -] \\
[- & + & + & + & + & - & + & + & + & +] \\
[+ & - & + & + & + & + & - & + & + & +] \\
[+ & + & - & + & + & + & + & - & + & +] \\
[+ & + & + & - & + & + & + & + & - & +] \\
[- & - & - & - & + & - & - & - & - & +] \\
[+ & - & - & - & - & + & - & - & - & -] \\
[- & + & - & - & - & - & + & - & - & -]
\end{matrix}$$

$$\begin{matrix}
[- & - & + & - & - & - & - & + & - & -] \\
[- & - & - & + & - & - & - & - & + & -] \\
[+ & - & + & - & + & - & + & - & + & -] \\
[- & + & - & + & - & + & - & + & - & +] \\
[+ & + & + & + & + & + & + & + & + & +] \\
[+ & + & - & - & - & + & + & - & - & -] \\
[- & + & + & - & - & - & + & + & - & -] \\
[- & - & + & + & - & - & - & + & + & -] \\
[- & - & - & + & + & - & - & - & + & +] \\
[+ & - & - & - & + & + & - & - & - & +] \\
[- & - & + & + & + & - & - & + & + & +] \\
[+ & - & - & + & + & + & - & - & + & +] \\
[+ & + & - & - & + & + & + & - & - & +] \\
[+ & + & + & - & - & + & + & + & - & -] \\
[- & + & + & + & - & - & + & + & + & -] \\
[- & - & - & - & - & - & - & - & - & -]
\end{matrix}$$

**11 ×11 circulant matrix**:

Let us consider an 11×11 matrix whose elements are 0, a, b, c, d, e, f, g, h, i, j which are represented in a circulant nature.

$$\begin{bmatrix}
0 & -4 & -3 & -2 & -1 & 6 & 6 & 1 & 2 & 3 & -4 \\
-4 & 0 & -4 & -3 & -2 & -1 & 6 & 6 & 1 & 2 & 3 \\
3 & -4 & 0 & -4 & -3 & -2 & -1 & 6 & 6 & 1 & 2 \\
2 & 3 & -4 & 0 & -4 & -3 & -2 & -1 & 6 & 6 & 1 \\
1 & 2 & 3 & -4 & 0 & -4 & -3 & -2 & -1 & 6 & 6 \\
6 & 1 & 2 & 3 & -4 & 0 & -4 & -3 & -2 & -1 & 6 \\
6 & 6 & 1 & 2 & 3 & -4 & 0 & -4 & -3 & -2 & -1 \\
-1 & 6 & 6 & 1 & 2 & 3 & -4 & 0 & -4 & -3 & -2 \\
-2 & -1 & 6 & 6 & 1 & 2 & 3 & -4 & 0 & -4 & -3 \\
-3 & -2 & -1 & 6 & 6 & 1 & 2 & 3 & -4 & 0 & -4 \\
-4 & -3 & -2 & -1 & 6 & 6 & 1 & 2 & 3 & -4 & 0
\end{bmatrix}$$

The above circulant matrix is 11×11 so the total outcomes of memory capacity are $2^{11}=2048$. The 11×11 matrix can hold 13 memories out of available memories and of those it can hold 11 unique memories, not counting the simple inversions.

$$\begin{matrix}
[+ & + & + & + & + & + & + & + & + & + & +] \\
[+ & + & + & + & + & - & + & + & + & + & -] \\
[- & + & + & + & + & + & - & + & + & + & +] \\
[+ & - & + & + & + & + & + & - & + & + & +] \\
[+ & + & - & + & + & + & + & + & - & + & +] \\
[+ & + & + & - & + & + & + & + & + & - & +] \\
[+ & + & + & + & - & + & + & + & + & + & -] \\
[- & + & + & + & + & - & + & + & + & + & +] \\
[+ & - & + & + & + & + & - & + & + & + & +] \\
[+ & + & - & + & + & + & + & - & + & + & +]
\end{matrix}$$

```
[+  +  +  −  +  +  +  +  −  +  +]
[+  +  +  +  −  +  +  +  +  −  +]
[−  −  −  −  −  −  −  −  −  −  −]
```

**12 ×12 circulant matrix**:

Let us consider a 12×12 matrix whose elements are 0, a, b, c, d, e, f, g, h, i, j, k which are represented in a circulant nature.

$$\begin{bmatrix}
0 & -5 & -4 & -1 & -2 & 3 & 17 & 2 & -1 & -3 & 4 & -5 \\
-5 & 0 & -5 & -4 & -1 & -2 & 3 & 17 & 2 & -1 & -3 & 4 \\
4 & -5 & 0 & -5 & -4 & -1 & -2 & 3 & 17 & 2 & -1 & -3 \\
-3 & 4 & -5 & 0 & -5 & -4 & -1 & -2 & 3 & 17 & 2 & -1 \\
-1 & -3 & 4 & -5 & 0 & -5 & -4 & -1 & -2 & 3 & 17 & 2 \\
2 & -1 & -3 & 4 & -5 & 0 & -5 & -4 & -1 & -2 & 3 & 17 \\
17 & 2 & -1 & -3 & 4 & -5 & 0 & -5 & -4 & -1 & -2 & 3 \\
3 & 17 & 2 & -1 & -3 & 4 & -5 & 0 & -5 & -4 & -1 & -2 \\
-2 & 3 & 17 & 2 & -1 & -3 & 4 & -5 & 0 & -5 & -4 & -1 \\
-1 & -2 & 3 & 17 & 2 & -1 & -3 & 4 & -5 & 0 & -5 & -4 \\
-4 & -1 & -2 & 3 & 17 & 2 & -1 & -3 & 4 & -5 & 0 & -5 \\
-5 & -4 & -1 & -2 & 3 & 17 & 2 & -1 & -3 & 4 & -5 & 0
\end{bmatrix}$$

The above circulant matrix is 12×12 so the total outcomes of memory capacity are $2^{12}$=4096. The 12×12 matrix can hold 34 memories out of available memories.

```
[+  +  +  +  +  +  +  +  +  +  +  +]
[+  +  +  +  +  −  +  +  +  +  +  −]
[−  −  −  −  −  +  −  −  −  −  −  +]
[−  +  +  +  +  +  −  +  +  +  +  +]
[+  −  −  −  −  −  +  −  −  −  −  −]
[+  −  +  +  +  +  +  −  +  +  +  +]
[−  +  −  −  −  −  −  +  −  −  −  −]
[+  +  −  +  +  +  +  +  −  +  +  +]
[−  −  +  −  −  −  −  −  +  −  −  −]
[+  +  +  −  +  +  +  +  +  −  +  +]
[−  −  −  +  −  −  −  −  −  +  −  −]
[+  +  +  +  −  +  +  +  +  +  −  +]
[−  −  −  −  +  −  −  −  −  −  +  −]
[+  −  +  −  +  −  +  −  +  −  +  −]
[−  +  −  +  −  +  −  +  −  +  −  +]
[+  +  −  −  −  −  +  +  −  −  −  −]
[−  −  +  +  +  +  −  −  +  +  +  +]
[−  +  +  −  −  −  −  +  +  −  −  −]
[+  −  −  +  +  +  +  −  −  +  +  +]
[−  −  +  +  −  −  −  −  +  +  −  −]
[+  +  −  −  +  +  +  +  −  −  +  +]
[−  −  −  +  +  −  −  −  −  +  +  −]
```

$$\begin{bmatrix}
+ & + & + & - & - & + & + & + & + & - & - & + \\
- & - & - & - & + & + & - & - & - & - & + & + \\
+ & + & + & + & - & - & + & + & + & + & - & - \\
+ & - & - & - & - & + & + & - & - & - & - & + \\
- & + & + & + & + & - & - & + & + & + & + & - \\
+ & + & - & + & + & - & + & + & - & + & + & - \\
- & - & + & - & - & + & - & - & + & - & - & + \\
- & + & + & - & + & + & - & + & + & - & + & + \\
+ & - & - & + & - & - & + & - & - & + & - & - \\
+ & - & + & + & - & + & + & - & + & + & - & + \\
- & + & - & - & + & - & - & + & - & - & + & - \\
- & - & - & - & - & - & - & - & - & - & - & -
\end{bmatrix}$$

**13 ×13 circulant matrix**:

Let us consider a 13×13 matrix whose elements are 0, a, b, c, d, e, f, g, h, i, j, k, l which are represented in a circulant nature

$$\begin{bmatrix}
0 & -5 & -4 & -3 & -2 & -1 & 8 & 8 & 1 & 2 & 3 & 4 & -5 \\
-5 & 0 & -5 & -4 & -3 & -2 & -1 & 8 & 8 & 1 & 2 & 3 & 4 \\
4 & -5 & 0 & -5 & -4 & -3 & -2 & -1 & 8 & 8 & 1 & 2 & 3 \\
3 & 4 & -5 & 0 & -5 & -4 & -3 & -2 & -1 & 8 & 8 & 1 & 2 \\
2 & 3 & 4 & -5 & 0 & -5 & -4 & -3 & -2 & -1 & 8 & 8 & 1 \\
1 & 2 & 3 & 4 & -5 & 0 & -5 & -4 & -3 & -2 & -1 & 8 & 8 \\
8 & 1 & 2 & 3 & 4 & -5 & 0 & -5 & -4 & -3 & -2 & -1 & 8 \\
8 & 8 & 1 & 2 & 3 & 4 & -5 & 0 & -5 & -4 & -3 & -2 & -1 \\
-1 & 8 & 8 & 1 & 2 & 3 & 4 & -5 & 0 & -5 & -4 & -3 & -2 \\
-2 & -1 & 8 & 8 & 1 & 2 & 3 & 4 & -5 & 0 & -5 & -4 & -3 \\
-3 & -2 & -1 & 8 & 8 & 1 & 2 & 3 & 4 & -5 & 0 & -5 & -4 \\
-4 & -3 & -2 & -1 & 8 & 8 & 1 & 2 & 3 & 4 & -5 & 0 & -5 \\
-5 & -4 & -3 & -2 & -1 & 8 & 8 & 1 & 2 & 3 & 4 & -5 & 0
\end{bmatrix}$$

The above circulant matrix is 13×13 so the total outcomes of memory capacity are $2^{13}$=8192. The 13×13 matrix can hold 15 memories out of available memories.

$$\begin{bmatrix}
+ & + & + & + & + & + & + & + & + & + & + & + & + \\
+ & + & + & + & + & + & - & + & + & + & + & + & - \\
- & + & + & + & + & + & + & - & + & + & + & + & + \\
+ & - & + & + & + & + & + & + & - & + & + & + & + \\
+ & + & - & + & + & + & + & + & + & - & + & + & + \\
+ & + & + & - & + & + & + & + & + & + & - & + & + \\
+ & + & + & + & - & + & + & + & + & + & + & - & + \\
+ & + & + & + & + & - & + & + & + & + & + & + & - \\
- & + & + & + & + & + & - & + & + & + & + & + & + \\
+ & - & + & + & + & + & + & - & + & + & + & + & + \\
+ & + & - & + & + & + & + & + & - & + & + & + & + \\
+ & + & + & - & + & + & + & + & + & - & + & + & + \\
+ & + & + & + & - & + & + & + & + & + & - & + & +
\end{bmatrix}$$

```
[+  +  +  +  +  −  +  +  +  +  +  −  +]
[−  −  −  −  −  −  −  −  −  −  −  −  −]
```

**14 ×14 circulant matrix**:

Let us consider a 14×14 matrix whose elements are 0, a, b, c, d, e, f, g, h, i, j, k, l, m which are represented in a circulant nature

$$\begin{bmatrix}
0 & -6 & -5 & -4 & -1 & -2 & 3 & 28 & 2 & -1 & 3 & -4 & 5 & -6 \\
-6 & 0 & -6 & -5 & -4 & -1 & -2 & 3 & 28 & 2 & -1 & 3 & -4 & 5 \\
5 & -6 & 0 & -6 & -5 & -4 & -1 & -2 & 3 & 28 & 2 & -1 & 3 & -4 \\
-4 & 5 & -6 & 0 & -6 & -5 & -4 & -1 & -2 & 3 & 28 & 2 & -1 & 3 \\
3 & -4 & 5 & -6 & 0 & -6 & -5 & -4 & -1 & -2 & 3 & 28 & 2 & -1 \\
-1 & 3 & -4 & 5 & -6 & 0 & -6 & -5 & -4 & -1 & -2 & 3 & 28 & 2 \\
2 & -1 & 3 & -4 & 5 & -6 & 0 & -6 & -5 & -4 & -1 & -2 & 3 & 28 \\
28 & 2 & -1 & 3 & -4 & 5 & -6 & 0 & -6 & -5 & -4 & -1 & -2 & 3 \\
3 & 28 & 2 & -1 & 3 & -4 & 5 & -6 & 0 & -6 & -5 & -4 & -1 & -2 \\
-2 & 3 & 28 & 2 & -1 & 3 & -4 & 5 & -6 & 0 & -6 & -5 & -4 & -1 \\
-1 & -2 & 3 & 28 & 2 & -1 & 3 & -4 & 5 & -6 & 0 & -6 & -5 & -4 \\
-4 & -1 & -2 & 3 & 28 & 2 & -1 & 3 & -4 & 5 & -6 & 0 & -6 & -5 \\
-5 & -4 & -1 & -2 & 3 & 28 & 2 & -1 & 3 & -4 & 5 & -6 & 0 & -6 \\
-6 & -5 & -4 & -1 & -2 & 3 & 28 & 2 & -1 & 3 & -4 & 5 & -6 & 0
\end{bmatrix}$$

The above circulant matrix is 14×14 so the total outcomes of memory capacity are $2^{14}=16384$. The 14×14 matrix can hold 46 memories out of available memories out of those 23 of them are unique.

```
[+  +  +  +  +  +  +  +  +  +  +  +  +  +]
[+  +  +  +  +  +  −  +  +  +  +  +  +  −]
[−  +  +  +  +  +  +  −  +  +  +  +  +  +]
[+  −  +  +  +  +  +  +  −  +  +  +  +  +]
[+  +  −  +  +  +  +  +  +  −  +  +  +  +]
[+  +  +  −  +  +  +  +  +  +  −  +  +  +]
[+  +  +  +  −  +  +  +  +  +  +  −  +  +]
[+  +  +  +  +  −  +  +  +  +  +  +  −  +]
[+  −  +  −  +  −  +  −  +  −  +  −  +  −]
[+  +  +  +  −  −  −  +  +  +  +  −  −  −]
[−  +  +  +  +  −  −  −  +  +  +  +  −  −]
[−  −  +  +  +  +  −  −  −  +  +  +  +  −]
[−  −  −  +  +  +  +  −  −  −  +  +  +  +]
[+  −  −  −  +  +  +  +  −  −  −  +  +  +]
[+  +  −  −  −  +  +  +  +  −  −  −  +  +]
[+  +  +  −  −  −  +  +  +  +  −  −  −  +]
[+  +  −  −  −  −  −  +  +  −  −  −  −  −]
[−  +  +  −  −  −  −  −  +  +  −  −  −  −]
[−  −  +  +  −  −  −  −  −  +  +  −  −  −]
[−  −  −  +  +  −  −  −  −  −  +  +  −  −]
```

```
[−  −  −  −  +  +  −  −  −  −  −  +  +  −]
[−  −  −  −  −  +  +  −  −  −  −  −  +  +]
[+  −  −  −  −  −  +  +  −  −  −  −  −  +]
```

**15 ×15 circulant matrix**:

Let us consider a 15×15 matrix whose elements are 0, a, b, c, d, e, f, g, h, i, j, k, l, m, n which are represented in a circulant nature

$$\begin{bmatrix} 0 & -6 & -5 & -4 & -3 & -2 & -1 & 10 & 10 & 1 & 2 & 3 & 4 & 5 & -6 \\ -6 & 0 & -6 & -5 & -4 & -3 & -2 & -1 & 10 & 10 & 1 & 2 & 3 & 4 & 5 \\ 5 & -6 & 0 & -6 & -5 & -4 & -3 & -2 & -1 & 10 & 10 & 1 & 2 & 3 & 4 \\ 4 & 5 & -6 & 0 & -6 & -5 & -4 & -3 & -2 & -1 & 10 & 10 & 1 & 2 & 3 \\ 3 & 4 & 5 & -6 & 0 & -6 & -5 & -4 & -3 & -2 & -1 & 10 & 10 & 1 & 2 \\ 2 & 3 & 4 & 5 & -6 & 0 & -6 & -5 & -4 & -3 & -2 & -1 & 10 & 10 & 1 \\ 1 & 2 & 3 & 4 & 5 & -6 & 0 & -6 & -5 & -4 & -3 & -2 & -1 & 10 & 10 \\ 10 & 1 & 2 & 3 & 4 & 5 & -6 & 0 & -6 & -5 & -4 & -3 & -2 & -1 & 10 \\ 10 & 10 & 1 & 2 & 3 & 4 & 5 & -6 & 0 & -6 & -5 & -4 & -3 & -2 & -1 \\ -1 & 10 & 10 & 1 & 2 & 3 & 4 & 5 & -6 & 0 & -6 & -5 & -4 & -3 & -2 \\ -2 & -1 & 10 & 10 & 1 & 2 & 3 & 4 & 5 & -6 & 0 & -6 & -5 & -4 & -3 \\ -3 & -2 & -1 & 10 & 10 & 1 & 2 & 3 & 4 & 5 & -6 & 0 & -6 & -5 & -4 \\ -4 & -3 & -2 & -1 & 10 & 10 & 1 & 2 & 3 & 4 & 5 & -6 & 0 & -6 & -5 \\ -5 & -4 & -3 & -2 & -1 & 10 & 10 & 1 & 2 & 3 & 4 & 5 & -6 & 0 & -6 \\ -6 & -5 & -4 & -3 & -2 & -1 & 10 & 10 & 1 & 2 & 3 & 4 & 5 & -6 & 0 \end{bmatrix}$$

The above circulant matrix is 15×15 so the total outcomes of memory capacity are $2^{15}$=32768. The 15×15 matrix can hold 17 memories out of available memories

```
[+  +  +  +  +  +  +  +  +  +  +  +  +  +  +]
[+  +  +  +  +  +  +  −  +  +  +  +  +  +  −]
[−  +  +  +  +  +  +  +  −  +  +  +  +  +  +]
[+  −  +  +  +  +  +  +  +  −  +  +  +  +  +]
[+  +  −  +  +  +  +  +  +  +  −  +  +  +  +]
[+  +  +  −  +  +  +  +  +  +  +  −  +  +  +]
[+  +  +  +  −  +  +  +  +  +  +  +  −  +  +]
[+  +  +  +  +  −  +  +  +  +  +  +  +  −  +]
[+  +  +  +  +  +  −  +  +  +  +  +  +  +  −]
[−  +  +  +  +  +  +  −  +  +  +  +  +  +  +]
[+  −  +  +  +  +  +  +  −  +  +  +  +  +  +]
[+  +  −  +  +  +  +  +  +  −  +  +  +  +  +]
[+  +  +  −  +  +  +  +  +  +  −  +  +  +  +]
[+  +  +  +  −  +  +  +  +  +  +  −  +  +  +]
[+  +  +  +  +  −  +  +  +  +  +  +  −  +  +]
[+  +  +  +  +  +  −  +  +  +  +  +  +  −  +]
[+  +  +  +  +  +  +  −  +  +  +  +  +  +  −]
[−  −  −  −  −  −  −  −  −  −  −  −  −  −  −]
```

**Experimental graphs**:
The above results when plotted on a graph give results that show increase in number of memories stored as the size increases. While plotting on a graph as first instance let us plot the memories held by the N×N circulant matrices and the graph generated will be as follows.

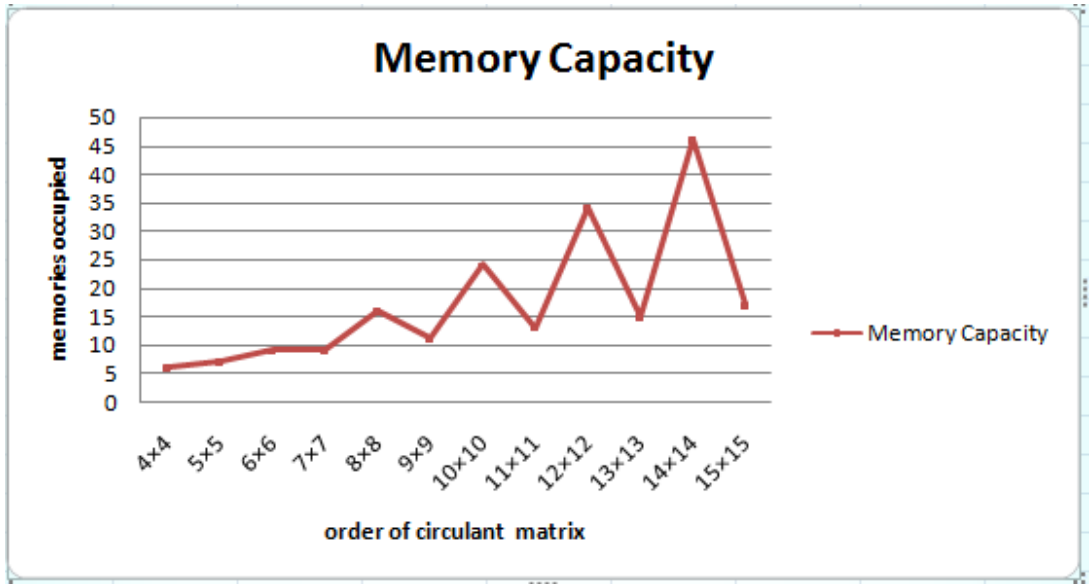

Figure 2: Memories occupied by each circulant matrix

As the circulant matrices of order 4×4, 6×6…14×14 shows some resemblance i.e. the sum of the elements add up to zero in these matrices and similarly circulant matrices 5×5, 7×7…15×5 shows some resemblance too.

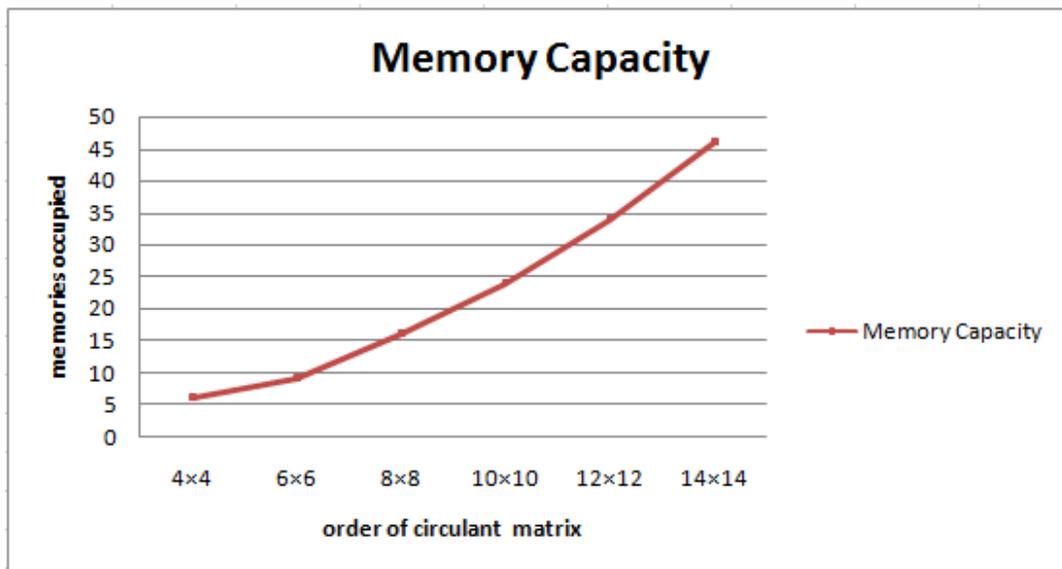

Figure 3: Memories occupied by even order circulant matrices up to 14×14

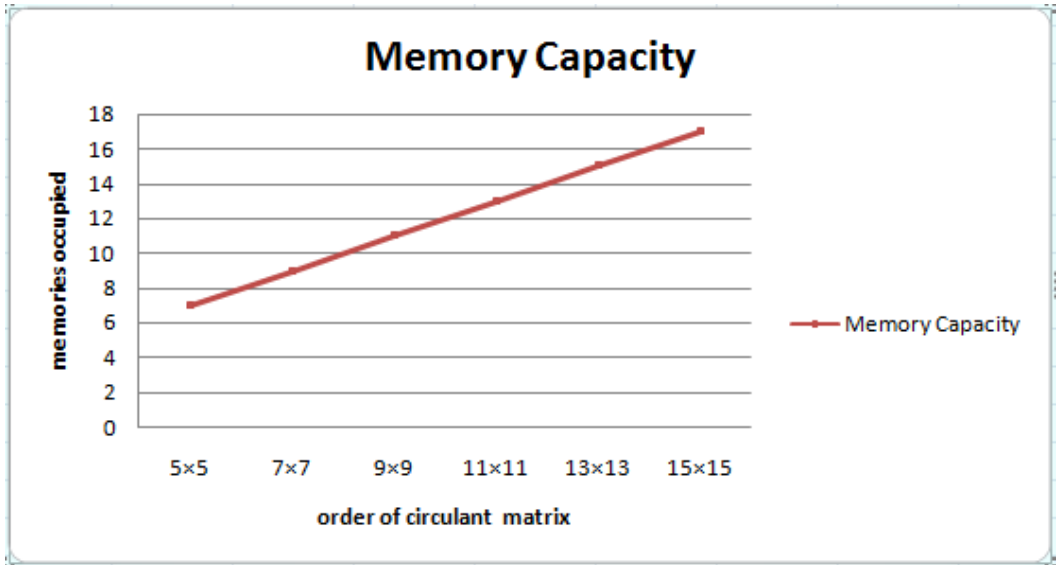

Figure 4: Memories occupied by odd order circulant matrices up to 15×15

The above graph explains the memories held by each of the matrix out of the total outcomes and it shows that how efficiently a circulant matrix can hold memories.

**Conclusions**

This paper presented that memories can be stored by using a circulant matrix instead of the weight matrix in traditional Hopfield network model. These are some general comments after doing experiments.

1. For the 4×4 case, the memories are basically ++ - - & its circular shifts.
2. For the 5×5 case, most of the memories are shifts of ++ - + - & +--++.
3. For the 6×6 case, the situation is more interesting. You have two classes;
   ++ - ++ - & + - + - + - and their complements.
4. For the 7×7 case, we have the all +, the all – sequences and shifts of
   +++ - ++ -.
5. For the 8×8 case, we have three classes +++-+++- , ++---++-- & +-+-+-+- and their complements.
6. For the 9×9 case, we have all the sequences as shifts of ++++-+++-.
7. For the 10×10 case, we have three classes ++++-++++- , ++----++---- & +-+-+-+-+- their shifts and complements.
8. For the 11×11 case, we have all the sequences as shifts of +++++-++++-.
9. For the 12×12 case, we have four classes +++++-+++++- , +-+-+-+-+-+- , ++------++-----, ++-++-++-++- their shifts and complements.
10. For the 13×13 case, we have all the sequences as shifts of ++++++-+++++-.
11. For the 14×14 case, we have four classes ++++++-+++++++- , +-+-+-+-+-+-+- , ++------++-----, ++++---++++---their shifts and complements.
12. For the 15×15 case, we have all the sequences as shifts of +++++++-++++++-.

Future research should explore applications of such structured weight networks to non-feedback networks [41]-[54]. It would be worthwhile to determine if there are general mathematical principles that make it possible to map certain visual structures associated with circulant and similar weight matrices.